\documentclass[letterpaper]{article}
\usepackage[preprint]{aaai2027}
\usepackage[hyphens]{url}
\usepackage{graphicx}
\usepackage{xcolor}
\usepackage{booktabs}
\usepackage{colortbl}
\usepackage{multirow}
\usepackage{subcaption}
\usepackage{natbib}
\usepackage{bm}
\usepackage{caption}
\usepackage{booktabs}
\usepackage{amsmath,amssymb}
\usepackage{colortbl}
\usepackage{xcolor}
\usepackage{makecell}
\usepackage[table]{xcolor}
\usepackage{stackengine}
\usepackage{adjustbox}

\title{CLEAR: Conflict-aware Learning via Evidence-guided Adaptive Routing \\ 
for Unified Sparse-View 3D Gaussian Super-Resolution}

\author{
Hantang Li\textsuperscript{1,2},
Qiang Zhu\textsuperscript{2},
Xiandong Meng\textsuperscript{2}\corresponding,\\
Debin Zhao\textsuperscript{3},
Xiaopeng Fan\textsuperscript{3,2}\corresponding
}

\affiliations{
\textsuperscript{1}Harbin Institute of Technology, Shenzhen, China\\
\textsuperscript{2}Pengcheng Laboratory, Shenzhen, China\\
\textsuperscript{3}Harbin Institute of Technology, Harbin, China\\
\texttt{25B951062@stu.hit.edu.cn},
\texttt{zhuqiang@pcl.ac.cn},
\texttt{mengxd@pcl.ac.cn}\\
\texttt{dbzhao@hit.edu.cn},
\texttt{fxp@hit.edu.cn}
}

\begin{document}
\maketitle

\begin{abstract}
Sparse-view 3D Gaussian Splatting Super-resolution is highly challenging since the sparse and low-resolution (LR) inputs lack sufficient geometric and high-frequency information for accurate reconstruction.
To achieve high-quality reconstruction, existing sparse-view super-resolution methods adhere to two-stage pipeline that performs LR Gaussian reconstruction and then high-resolution (HR) Gaussian refinement, which directly results in stage-wise Gaussian transfer and reconstruction error accumulation. To this end, we propose \textbf{CLEAR}, a \textbf{C}onflict-aware \textbf{L}earning via \textbf{E}vidence-guided \textbf{A}daptive \textbf{R}outing, as the first unified single-stage framework for Sparse-view 3D Gaussian Splatting Super-resolution. Specifically, CLEAR performs joint the optimization of authentic LR observations and external HR priors within a unified  Gaussian representation.    
To mitigate the gradient conflicts introduced by sparse supervision during training, we propose a Gaussian-wise conflict-aware optimization strategy that regards the LR gradient as a reliable anchor and applies evidence-conditioned soft correction only to severe HR conflicts.
Moreover, to recover high-frequency details, we introduce an evidence-guided Patch-to-Gaussian routing mechanism which estimates patch reliability and detail demand, lifts them into Gaussian space, and selectively routes high-frequency gradients and densification. Finally, we employ shared Gaussian dropout and a detached mid-training anchoring to enhance the robustness of training framework.   Extensive experiments on both synthetic and real-world $4\times$ super-resolution benchmarks demonstrate that CLEAR consistently achieves state-of-the-art rendering quality and superior geometric fidelity.

\end{abstract}


\section{Introduction}
\label{sec:introduction}

Sparse-view 3D Gaussian super-resolution aims to reconstruct high-quality, high-resolution (HR) 3D scenes from sparse and low-resolution (LR) view inputs, which is essential for applications such as virtual/augmented reality, robotics, and digital twins. However, the lack of sufficient geometric and high-frequency information makes this task highly challenging. 
On one  hand, existing 3DGS super-resolution methods hinge on dense LR views ~\cite{srgs,gaussiansr,supergaussian,splatsure} for HR scene reconstruction, yet suffer from geometric ambiguity and loss of high-frequency cues under sparse views.
On the other hand, sparse-view 3D scene reconstruction methods enhance geometric consistency via depth priors, frequency regularization or geometry-aware optimization, while they fail to recover high-frequency details~\cite{regnerf,sparsenerf,fsgs,corgs,coadaptation,dropoutgs,docgs,pairdropgs}. 
Consequently, sparse-view 3D Gaussian super-resolution remains a formidable challenge, as geometry and detail recovery are inherently coupled and mutually dependent.
As illustrated in Fig.~\ref{fig:motivation}(a), existing sparse-view 3D Gaussian super-resolution frameworks~\cite{s2gaussian}, adopt a two-stage pipeline that first reconstructs an LR Gaussian field and then refines a transferred HR Gaussian field. While effective, this stage-wise paradigm increases pipeline complexity, prevents end-to-end optimization, and permits errors from LR reconstruction to cascade into the subsequent HR refinement stage.


To address these limitations, we propose CLEAR, to the best of our knowledge, the first unified single-stage framework for sparse-view 3D Gaussian super-resolution.  
However, unifying LR and HR supervision within a single optimization introduces a new challenge, i.e., gradient conflict during training. To analysis this issue, we first define the conflict ratio that denotes the proportion of Gaussians whose LR and HR gradients point in opposing directions, averaged over the training process. 
As illustrated in Fig.~\ref{fig:motivation}(b), Gaussian-wise LR and HR gradient conflicts become increasingly severe as the number of input views decreases. This observation indicates that sparse-view settings substantially aggravate the optimization conflicts between authentic LR supervision and external HR priors, making unified single-stage optimization considerably more challenging. 
Based on this observation, we develop a Gaussian-wise conflict-aware
optimization strategy that treats authentic LR gradients as reliable anchors
and applies evidence-conditioned soft correction only to severe HR conflicts.
Moreover, to recover the reliable high-frequency information, we propose an evidence-guided Patch-to-Gaussian routing mechanism to selectively inject trustworthy details and guide Gaussian densification. 
Since sparse-view optimization is prone to overfitting due to limited multi-view supervision, we further develop a stabilization strategy to enhance geometric consistency and generalization. In summary, our main contributions are as follows:

\begin{figure*}[!t]
\centering

\begin{subfigure}[t]{0.4\linewidth}
    \centering
    \includegraphics[width=\linewidth]{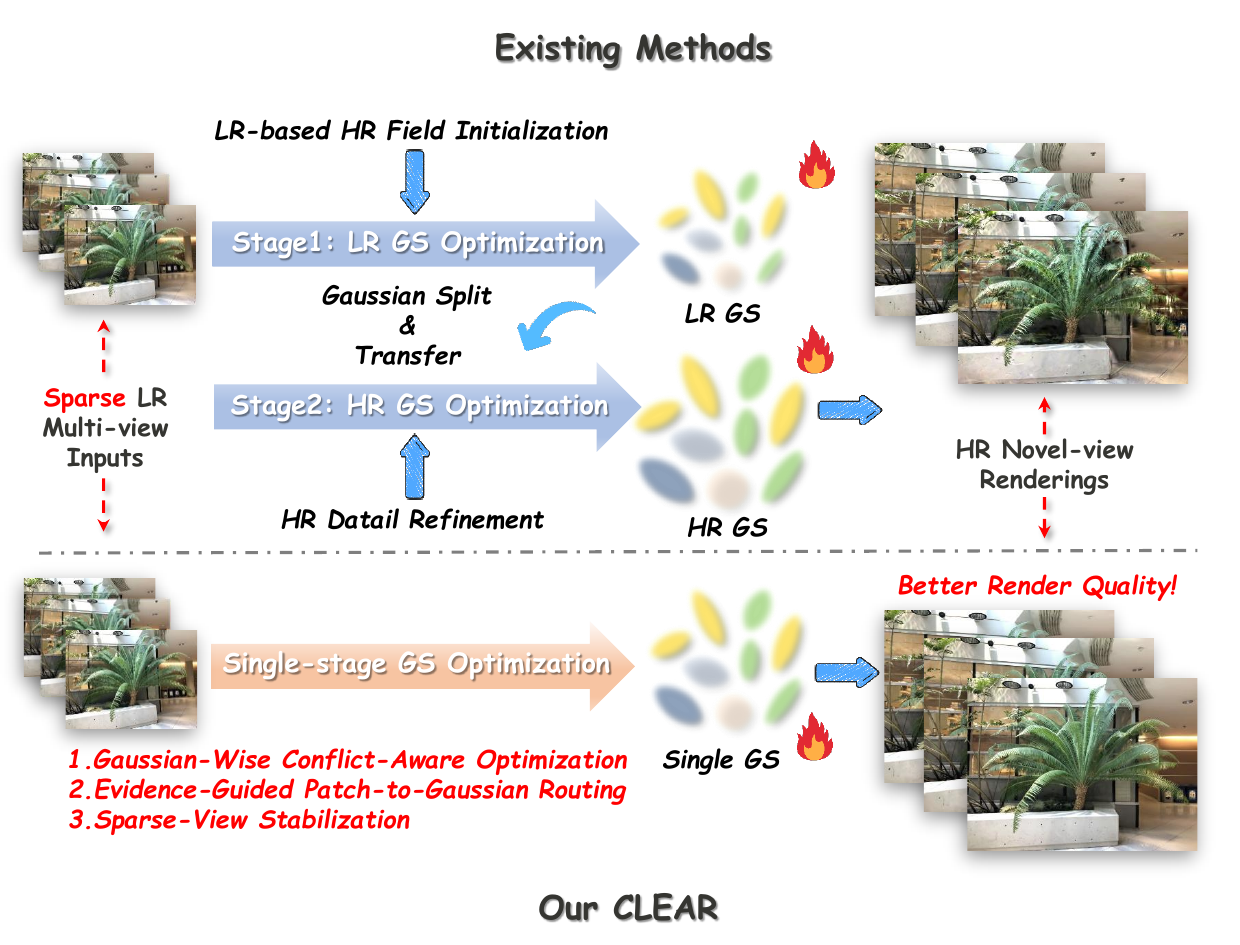}
    \par
    \vspace{-3pt}
    {\footnotesize (a)}
\end{subfigure}
\hfill
\begin{subfigure}[t]{0.3\linewidth}
    \centering
    \includegraphics[width=\linewidth]{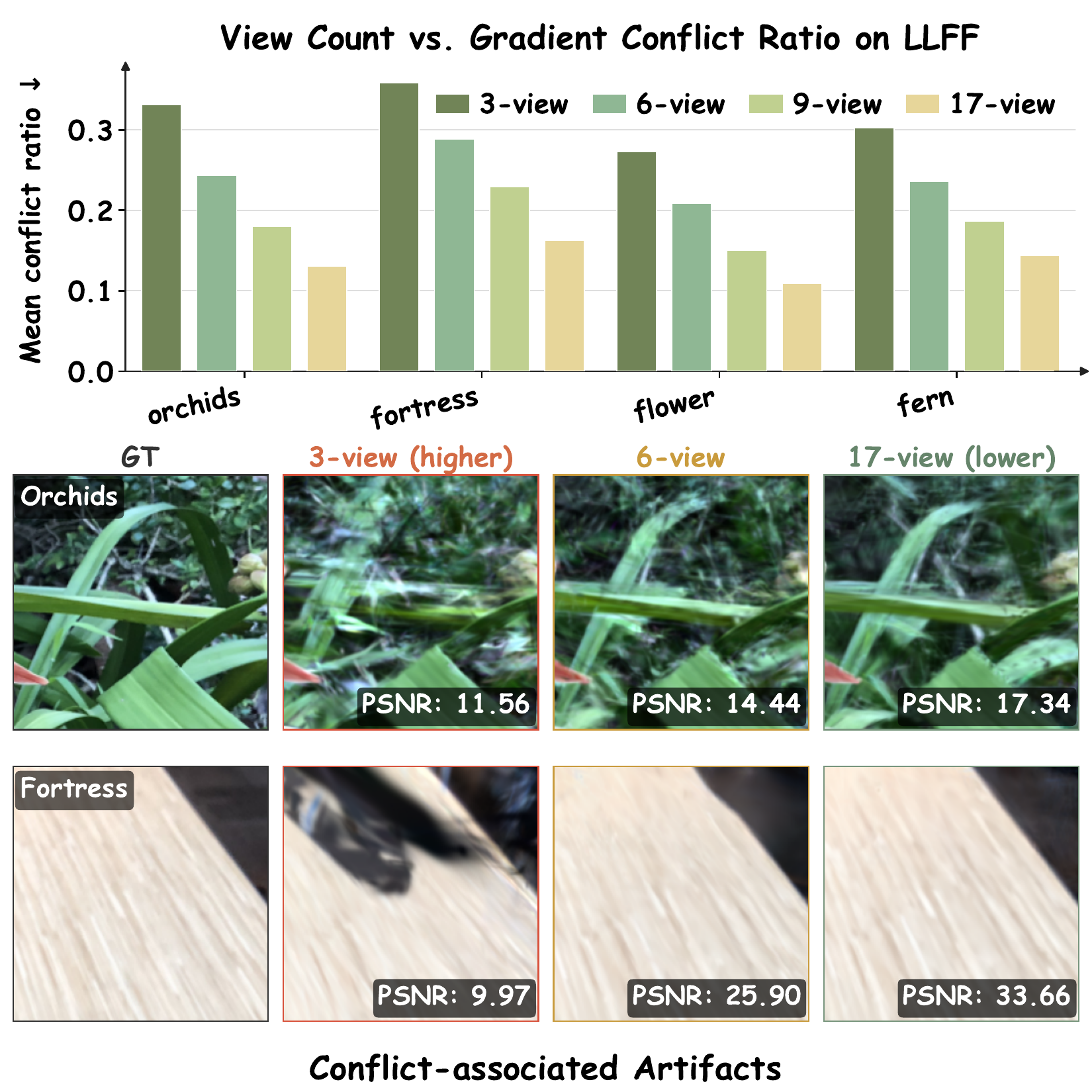}
    \par
    \vspace{-3pt}
    {\footnotesize (b)}
\end{subfigure}
\hfill
\begin{subfigure}[t]{0.29\linewidth}
    \centering
    \includegraphics[width=\linewidth]{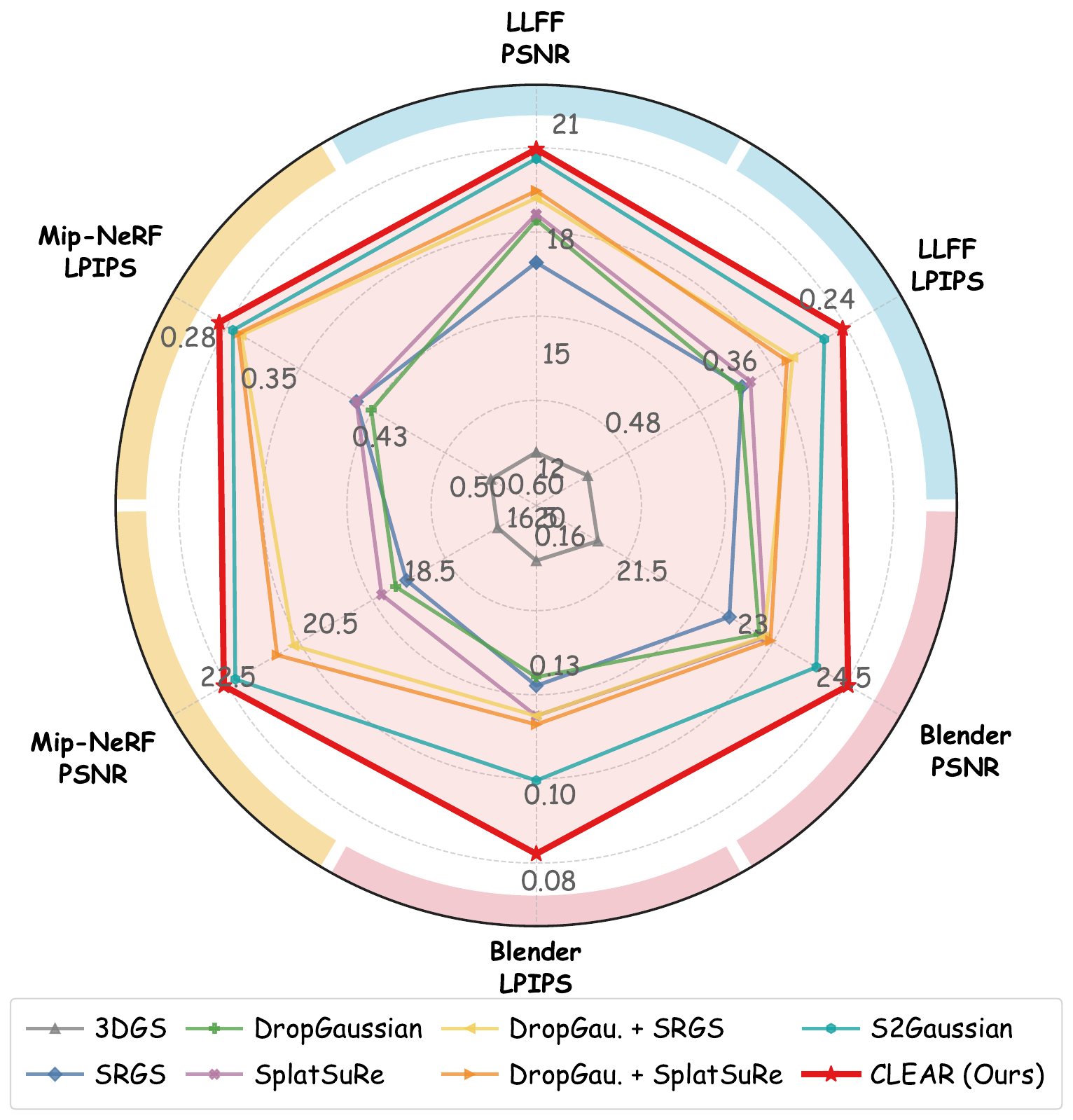}
    \par
    \vspace{-3pt}
    {\footnotesize (c)}
\end{subfigure}
\setlength{\abovecaptionskip}{-0.01pt} 
\caption{
\textbf{Motivation and Performance.}
(a) Unlike existing two-stage methods, CLEAR is the first unified single-stage framework that jointly optimizes LR and HR supervision within a single Gaussian field.
(b) Sparse-view observations intensify Gaussian-wise LR and HR gradient conflicts, leading to severe novel-view artifacts.
(c) The radar plot summarizes PSNR and LPIPS across LLFF, Blender, and Mip-NeRF 360, where CLEAR achieves consistent improvements across different scenes.
}
\label{fig:motivation}
\end{figure*}

\begin{itemize}

\item We propose \textbf{CLEAR}, to our knowledge, the first unified single-stage framework for sparse-view 3D Gaussian super-resolution. CLEAR jointly optimizes a unified Gaussian field under authentic LR observations and external HR supervision throughout training, avoiding separate LR/HR representations and stage-wise transfer.

\item We reveal that LR and HR gradient conflicts are particularly severe under sparse-view settings and introduce a Gaussian-wise conflict-aware optimization strategy that uses authentic LR gradients as anchors to suppress destructive HR updates while preserving beneficial cross-resolution corrections.

\item We develop an evidence-guided Patch-to-Gaussian routing mechanism that lifts patch-level SR reliability into Gaussian field and selectively guides high-frequency learning and Gaussian densification, thereby achieving reliable detail reconstruction. Moreover, we introduce sparse-view stabilization through shared Gaussian dropout and detached mid-training anchoring.

\end{itemize}

Extensive experiments on both synthetic and real-world $4\times$ super-resolution benchmarks demonstrate that CLEAR consistently achieves state-of-the-art rendering quality with improved geometric fidelity and perceptual realism, as shown in Fig.~\ref{fig:motivation}(c). 

\section{Related Work}

\subsection{Novel View Synthesis}

Novel view synthesis aims to reconstruct a scene representation from captured images and render images from unseen viewpoints. Neural Radiance Fields (NeRF) achieves high-quality synthesis via implicit neural functions and volume endering~\cite{nerf,mipnerf,mipnerf360}. Subsequent works accelerate this process using explicit feature grids, tensor decomposition, or hash encoding~\cite{tensorf,plenoxels}.
Recently, 3D Gaussian Splatting explicit anisotropic Gaussians and differentiable rasterization for fast optimization and real-time rendering~\cite{3dgs}. Follow-up methods boost quality and scalability via anti-aliasing, structured Gaussians, and progressive propagation~\cite{mipsplatting, scaffoldgs}. Despite their strong performance, these methods typically require dense, high-quality multi-view observations.

\subsection{Sparse Novel View Synthesis}

Sparse novel view synthesis reconstructs 3D scenes from only a few inputs. NeRF-based methods mitigate insufficient supervision via depth priors, frequency regularization, virtual-view constraints, or pretrained geometry ~\cite{depthnerf,regnerf,freenerf,sparsenerf}. Recent 3DGS-based methods address sparse-view overfitting through depth regularization, improved initialization, structural constraints, and Gaussian regularization ~\cite{li2024dngaussian,fsgs,corgs,sparsegs}. DropGaussian ~\cite{dropgaussian} and DropoutGS ~\cite{dropoutgs} randomly drop Gaussians during training to reduce overfitting; NexusGS ~\cite{nexusgs} and CoMapGS ~\cite{comapgs} improve geometry via epipolar depth priors and covisibility. Others incorporate self-ensembling or generative priors to compensate for missing observations ~\cite{segs,gsgs,oraclegs}. Despite improving geometry, none of these methods recover high-frequency details lost in low-resolution inputs.

\begin{figure*}[t]
    \centering
    \includegraphics[width=0.95\linewidth]{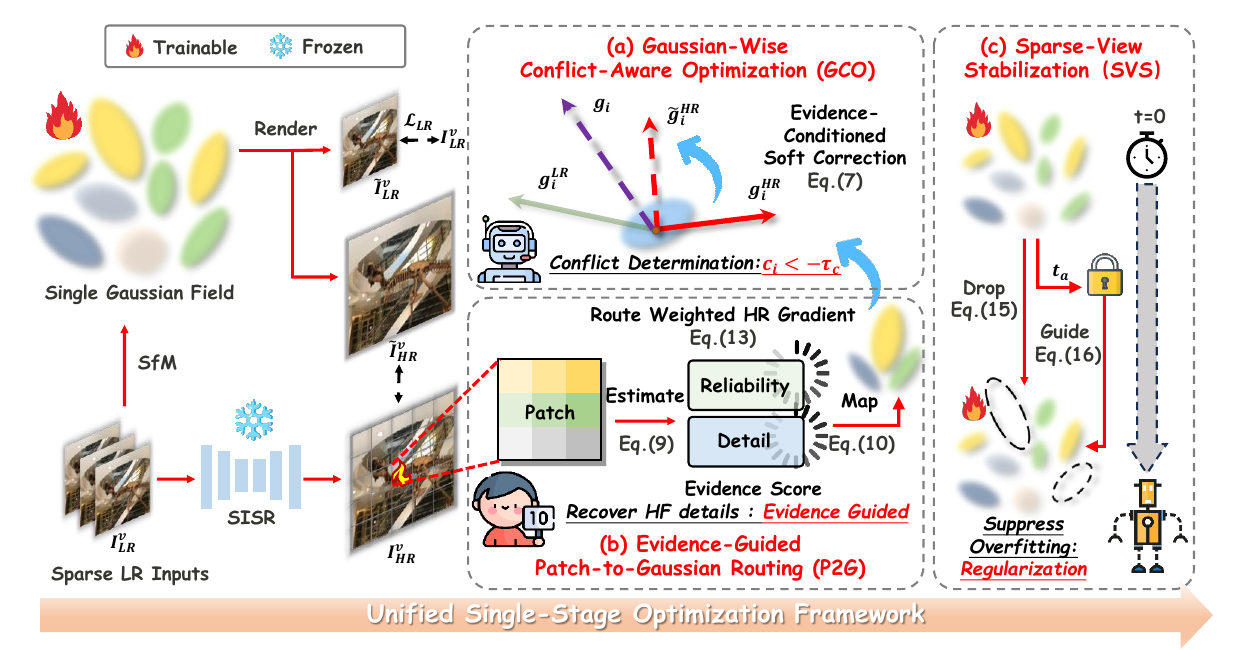}
    \caption{
        \textbf{Overview of the proposed CLEAR framework.} Given sparse LR observations, CLEAR optimizes a Gaussian field along a unified single-stage optimization, where (a) Gaussian-wise Conflict-aware Optimization reconciles LR and HR gradient conflicts, (b) Evidence-guided Patch-to-Gaussian Routing transfers reliable detail cues from SR patches to visible Gaussians, and (c) Sparse-view Stabilization regularizes the Gaussian field for robust HR novel-view rendering.
        }
    \label{fig:framework}
\end{figure*}

\subsection{Super-Resolution Novel View Synthesis}

Super-resolution novel view synthesis reconstructs HR 3D representations from LR multi-view images. Early NeRF-based methods recover fine details via supersampling, high-resolution references, multi-view image SR, or pretrained diffusion models ~\cite{wang2022nerfsr,huang2023refsr,supernerf}. Recent 3DGS-based methods leverage explicit Gaussians for efficient HR rendering: SRGS injects textures from a pretrained image SR model, GaussianSR uses diffusion priors, and SuperGaussian transfers temporal information from video SR models ~\cite{srgs,gaussiansr,supergaussian}. Later works enhance multi-view consistency via explicit 3D representations, uncertainty modeling, or selective detail enhancement ~\cite{3dsr, splatsure}. S2Gaussian ~\cite{s2gaussian} employs a two-stage pipeline that transfers an LR Gaussian representation to initialize HR refinement. In contrast, CLEAR performs sparse-view 3D Gaussian super-resolution in a unified single-stage optimization, maintaining a single Gaussian field throughout training while jointly exploiting LR observations and external HR priors—without stage-wise transfer.

\section{Methodology}
\label{sec:method}

\subsection{Preliminaries}
\label{sec:preliminaries}

\paragraph{3D Gaussian Splatting (3DGS)} represents a scene as a collection of anisotropic
Gaussian primitives
$\mathcal{G}_{\theta}=\{\mathcal{G}_i\}_{i=1}^{G}$.
Each Gaussian is formulated as:
\begin{equation}
g_i(\mathbf{x})
=
\exp\left(
-\frac{1}{2}
(\mathbf{x}-\boldsymbol{\mu}_i)^\top
\boldsymbol{\Sigma}_i^{-1}
(\mathbf{x}-\boldsymbol{\mu}_i)
\right),
\label{eq:gaussian}
\end{equation}
where $g_i(\mathbf{x})$ denotes the density value of the $i$-th Gaussian
at the 3D location $\mathbf{x}$,
$\mathbf{x}\in\mathbb{R}^{3}$ denotes a 3D point,
$\boldsymbol{\mu}_i\in\mathbb{R}^{3}$ is the Gaussian center, and
$\boldsymbol{\Sigma}_i\in\mathbb{R}^{3\times3}$ is the covariance matrix, $\theta$
denotes all learnable Gaussian parameters. 
During optimization, adaptive density control dynamically clones, splits, and
prunes Gaussian primitives to refine the scene representation.
CLEAR builds upon 3DGS by introducing an additive view-independent detail
feature, scale-aware Mip rasterization, and a unified optimization strategy
for sparse-view 3D Gaussian super-resolution.
\subsection{Unified Single-Stage Framework (CLEAR)}
\label{sec:overview}

We formulate sparse-view 3D Gaussian super-resolution as a unified
single-stage optimization problem. Given sparse LR observations and their
camera parameters
$\{I_{\mathrm{LR}}^{v},C^v\}_{v=1}^{V}$, we generate external SR references
$I_{\mathrm{HR}}^{v}$ using a frozen image SR model, where $v$ indexes the
view, $V$ is the number of input views, and $C^v$ denotes the camera
parameters. Unlike previous two-stage method~\cite{s2gaussian}, CLEAR maintains a single Gaussian field $\mathcal G_{\theta}$ throughout optimization. An overview of the proposed framework is illustrated in
Fig.~\ref{fig:framework}.

For each view, the same Gaussian field is rendered at the LR and HR scales:
\begin{equation}
\begin{aligned}
\widehat I_{\mathrm{LR}}^{v}
&=
\mathcal R_{\mathrm{LR}}
\left(\mathcal G_{\theta},C^v\right),\\
\widehat I_{\mathrm{HR}}^{v}
&=
\mathcal R_{\mathrm{HR}}
\left(\mathcal G_{\theta},C^v\right),
\end{aligned}
\label{eq:dual_rendering}
\end{equation}
where $\widehat I_{\mathrm{LR}}^{v}$ and
$\widehat I_{\mathrm{HR}}^{v}$ are the rendered images, and
$\mathcal R_{\mathrm{LR}}$ and $\mathcal R_{\mathrm{HR}}$ denote
scale-aware Mip rasterization at the corresponding resolutions.
While both scales share the same Gaussian geometry, HR rendering requires
additional capacity for fine appearance. We therefore define:
\begin{equation}
\bm f_i^{\mathrm{LR}}
=
\bm f_i^{\mathrm{base}},
\qquad
\bm f_i^{\mathrm{HR}}
=
\bm f_i^{\mathrm{base}}+\bm f_i^{\mathrm{det}},
\label{eq:additive_detail_feature}
\end{equation}
where $i$ indexes a Gaussian,
$\bm f_i^{\mathrm{LR}}$ and $\bm f_i^{\mathrm{HR}}$ are its appearance
features for LR and HR rendering,
$\bm f_i^{\mathrm{base}}$ is its shared base appearance feature, and
$\bm f_i^{\mathrm{det}}$ is its view-independent HR detail residual. The authentic LR supervision is:
\begin{equation}
\begin{aligned}
\mathcal L_{\mathrm{LR}}
={}&
(1-\lambda_{\mathrm{ssim}})
\left\|
\widehat I_{\mathrm{LR}}^{v}-I_{\mathrm{LR}}^{v}
\right\|_1\\
&+
\lambda_{\mathrm{ssim}}
\left[
1-\operatorname{SSIM}
\left(
\widehat I_{\mathrm{LR}}^{v},I_{\mathrm{LR}}^{v}
\right)
\right],
\end{aligned}
\label{eq:lr_objective}
\end{equation}
where $\lambda_{\mathrm{ssim}}$ balances the LR photometric and structural
terms. Accordingly, the external HR supervision consists of two complementary
objectives:
\begin{equation}
\begin{aligned}
\mathcal L_{\mathrm{base}}
={}&
(1-\lambda_{\mathrm{ssim}}^{\mathrm{HR}})
\left\|
\widehat I_{\mathrm{HR}}^{v}
-
I_{\mathrm{HR}}^{v}
\right\|_1\\
&+
\lambda_{\mathrm{ssim}}^{\mathrm{HR}}
\left[
1-
\operatorname{SSIM}
\left(
\widehat I_{\mathrm{HR}}^{v},
I_{\mathrm{HR}}^{v}
\right)
\right],\\
\mathcal L_{\mathrm{HF}}
={}&
\mathcal L_{\mathrm{Haar}}
\left(
\widehat I_{\mathrm{HR}}^{v},
I_{\mathrm{HR}}^{v}
\right),
\end{aligned}
\label{eq:hr_supervision}
\end{equation}
where $\mathcal L_{\mathrm{base}}$ combines the HR photometric and structural
terms, $\lambda_{\mathrm{ssim}}^{\mathrm{HR}}$ balances these two terms, $\mathcal L_{\mathrm{Haar}}$ denotes the Haar-domain reconstruction loss
computed over the high-frequency subbands, and
$\mathcal L_{\mathrm{HF}}$ is the resulting high-frequency supervision term. These objectives provide complementary base-appearance
and high-frequency supervision. Together, the authentic LR observations
preserve the scene structure, while the external HR references enrich the same
Gaussian field with fine appearance details.

\subsection{Gaussian-Wise Conflict-Aware Optimization}
\label{sec:conflict}

As discussed above, joint LR and HR supervision may induce conflicting updates on
the same Gaussian under sparse-view observations. We therefore perform
conflict-aware optimization at the Gaussian level. For Gaussian $i$, we compare
the LR gradient with the HR gradient:
\begin{equation}
\begin{aligned}
\bm g_i^{\mathrm{LR}}
&=
\nabla_{\theta_i}\mathcal L_{\mathrm{LR}},\\
c_i
&=
\frac{
\left\langle
\bm g_i^{\mathrm{LR}},
\bm g_i^{\mathrm{HR}}
\right\rangle
}{
\|\bm g_i^{\mathrm{LR}}\|_2
\|\bm g_i^{\mathrm{HR}}\|_2+\epsilon
},
\end{aligned}
\label{eq:gaussian_conflict}
\end{equation}
where $\theta_i$ denotes the  attributes of Gaussian $i$,
$\bm g_i^{\mathrm{LR}}$ is the authentic LR gradient,
$\bm g_i^{\mathrm{HR}}$ denotes the HR gradient supplied to the conflict-correction operator, $c_i$ is their cosine similarity, and $\epsilon$ ensures numerical stability. The cosine is computed independently for position, base appearance, higher-order appearance, opacity, scale, and rotation, with the group index omitted for clarity. A strong conflict is detected when $c_i <-\tau_c$, where $\tau_c\in[0,1)$ is the tolerance threshold.

Since conflicting HR gradients may still contain useful detail cues, completely discarding them can lead to overly conservative optimization. We instead apply an evidence-conditioned soft correction:
\begin{equation}
\begin{aligned}
d_i
={}&
\left\langle
\bm g_i^{\mathrm{LR}},
\bm g_i^{\mathrm{HR}}
\right\rangle+\tau_c
\|\bm g_i^{\mathrm{LR}}\|_2
\|\bm g_i^{\mathrm{HR}}\|_2,\\
\widetilde{\bm g}_i^{\mathrm{HR}}
={}&
\bm g_i^{\mathrm{HR}}-
\omega_i\,
\mathbb I[c_i<-\tau_c]
\frac{d_i}{
\|\bm g_i^{\mathrm{LR}}\|_2^2+\epsilon
}
\bm g_i^{\mathrm{LR}},
\end{aligned}
\label{eq:soft_conflict_correction}
\end{equation}
where $d_i$ is the signed violation of the compatibility margin, $\mathbb I[\cdot]$ is the indicator function, $\widetilde{\bm g}_i^{\mathrm{HR}}$ is the corrected HR gradient, and
$\omega_i\in[0,1]$ is the adaptive correction weight.

The conflict-corrected supervision gradient is:
\begin{equation}
\bm g_i
=
\bm g_i^{\mathrm{LR}}
+
\lambda_{\mathrm{HR}}
\widetilde{\bm g}_i^{\mathrm{HR}},
\label{eq:conflict_merged_gradient}
\end{equation}
where $\bm g_i$ is the conflict-corrected supervision gradient and
$\lambda_{\mathrm{HR}}$ balances the LR and HR supervision. By performing Gaussian-wise conflict-aware optimization, CLEAR preserves LR-supported structure while suppressing destructive HR updates, providing a stable basis for reliable detail enhancement.

\subsection{Evidence-Guided Patch-to-Gaussian Routing}
\label{sec:routing}

Conflict correction suppresses unreliable HR updates but cannot determine
where additional details should be learned. We estimate the reliability and
detail demand of each HR patch $p$ as:
\begin{equation}
\begin{aligned}
R_p
&=
\exp\left(
-\frac{
\operatorname{Avg}_{p}
\left|
\mathcal D(I_{\mathrm{HR}}^{v})-I_{\mathrm{LR}}^{v}
\right|
}{\tau_r}
\right),\\
A_p
&=
\left[
\frac{
\operatorname{Avg}_{p}
\left|
I_{\mathrm{HR}}^{v}
-\mathcal U(\mathcal D(I_{\mathrm{HR}}^{v}))
\right|
}{
2 \overline e^{\mathrm{det}}+\epsilon
}
\right]_{0}^{1},
\end{aligned}
\label{eq:patch_evidence}
\end{equation}
where $\mathcal D(\cdot)$ and $\mathcal U(\cdot)$ denote downsampling and upsampling operations,
$\operatorname{Avg}_{p}$ denotes averaging within patch $p$,
$\tau_r$ controls reliability sensitivity,
$\overline e^{\mathrm{det}}$ denotes the image-level mean detail residual, and
$[\cdot]_0^1$ denotes clipping to $[0,1]$.
Thus, $R_p$ measures LR consistency, while $A_p$ measures high-frequency
content.

For each visible Gaussian $i$, the patch evidence is lifted into Gaussian
field:
\begin{equation}
q_i
=
R_{p(\pi(\boldsymbol{\mu}_i))},
\qquad
a_i
=
A_{p(\pi(\boldsymbol{\mu}_i))},
\qquad
w_i=q_i a_i,
\label{eq:evidence_lifting}
\end{equation}
where $\boldsymbol{\mu}_i$ is the Gaussian center, $\pi(\cdot)$ denotes camera
projection, $p(\cdot)$ selects the corresponding patch, $q_i$ denotes
reliability, $a_i$ denotes detail demand, and $w_i$ denotes their joint
evidence. The reliability controls conflict correction through:
\begin{equation}
s_i
=
\left[
\frac{-c_i-\tau_c}{1-\tau_c}
\right]_{0}^{1},
\qquad
\omega_i=s_i(1-q_i),
\label{eq:evidence_conflict_weight}
\end{equation}
where $s_i$ is the conflict severity and $\omega_i$ is the correction weight.
Then, we define the detail-routing weight as:
\begin{equation}
m_i=q_{\min}+(1-q_{\min})w_i,
\label{eq:detail_routing_weight}
\end{equation}
where $m_i$ is the routing weight and $q_{\min}$ preserves minimum detail
supervision.

The HR gradient supplied to the conflict-correction operator is instantiated
as:
\begin{equation}
\bm g_i^{\mathrm{HR}}
=
\underbrace{\nabla_{\theta_i}\mathcal L_{\mathrm{base}}}
_{\bm g_i^{\mathrm{base}}}
+
\lambda_{\mathrm{HF}}m_i
\underbrace{\nabla_{\theta_i}\mathcal L_{\mathrm{HF}}}
_{\bm g_i^{\mathrm{HF}}},
\label{eq:routed_hr_gradient}
\end{equation}
where $\bm g_i^{\mathrm{base}}$ and $\bm g_i^{\mathrm{HF}}$ are the gradients
of $\mathcal L_{\mathrm{base}}$ and $\mathcal L_{\mathrm{HF}}$, respectively.
The detail feature $\bm f_i^{\mathrm{det}}$ is excluded from conflict
correction and updated only by the routed HR detail gradient.

The same evidence guides densification:
\begin{equation}
S_i^{\mathrm{dens}}
=
w_i
\left\|
\nabla_{\bm u_i}\mathcal L_{\mathrm{HF}}
\right\|_2,
\label{eq:evidence_densification}
\end{equation}
where $S_i^{\mathrm{dens}}$ is the densification score and
$\bm u_i=\pi(\boldsymbol{\mu}_i)$ is the projected Gaussian center. The
screen-space gradient is used only for densification. This routing design enables selective high-frequency learning and evidence-guided densification, improving fine-detail reconstruction while suppressing unreliable SR artifacts.

\begin{figure*}[t]
\vspace{-5pt}
\centering
\setlength{\tabcolsep}{1pt}

\begin{tabular}{ccccccc}

\includegraphics[width=0.168\textwidth]{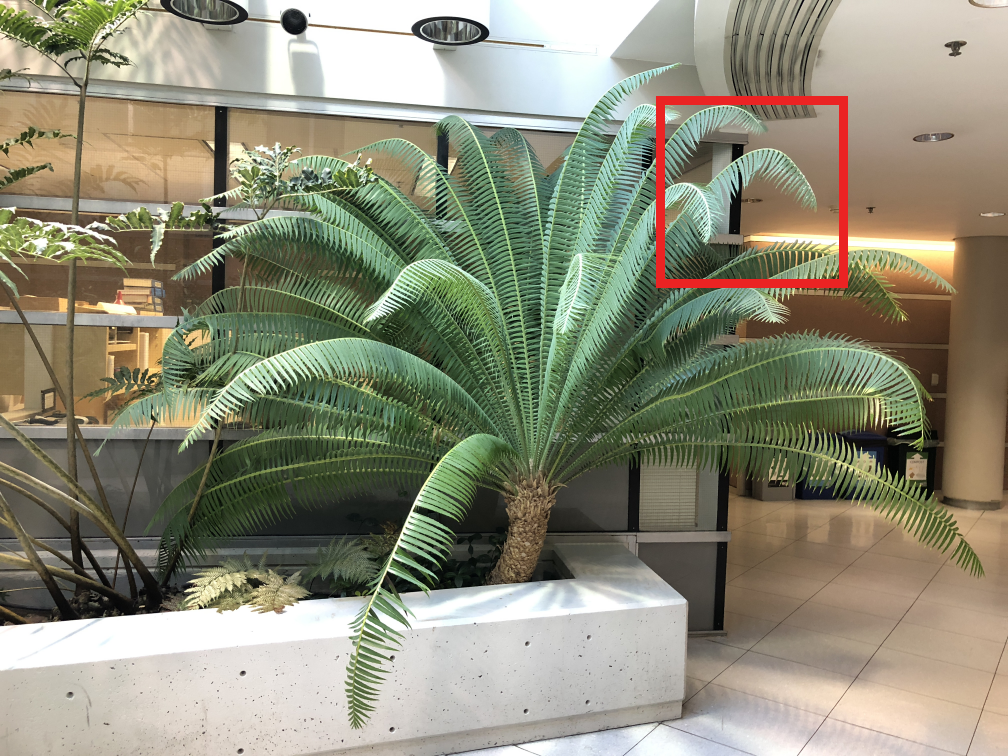} &
\includegraphics[width=0.127\textwidth]{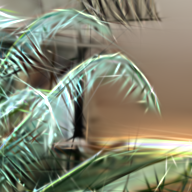} &
\includegraphics[width=0.127\textwidth]{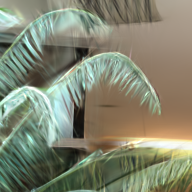} &
\includegraphics[width=0.127\textwidth]{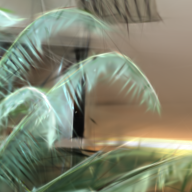} &
\includegraphics[width=0.127\textwidth]{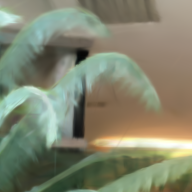} &
\includegraphics[width=0.127\textwidth]{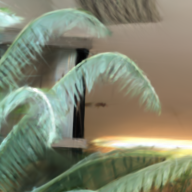} &
\includegraphics[width=0.127\textwidth]{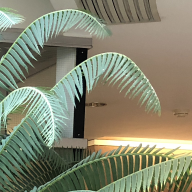}
\\

\includegraphics[width=0.168\textwidth]{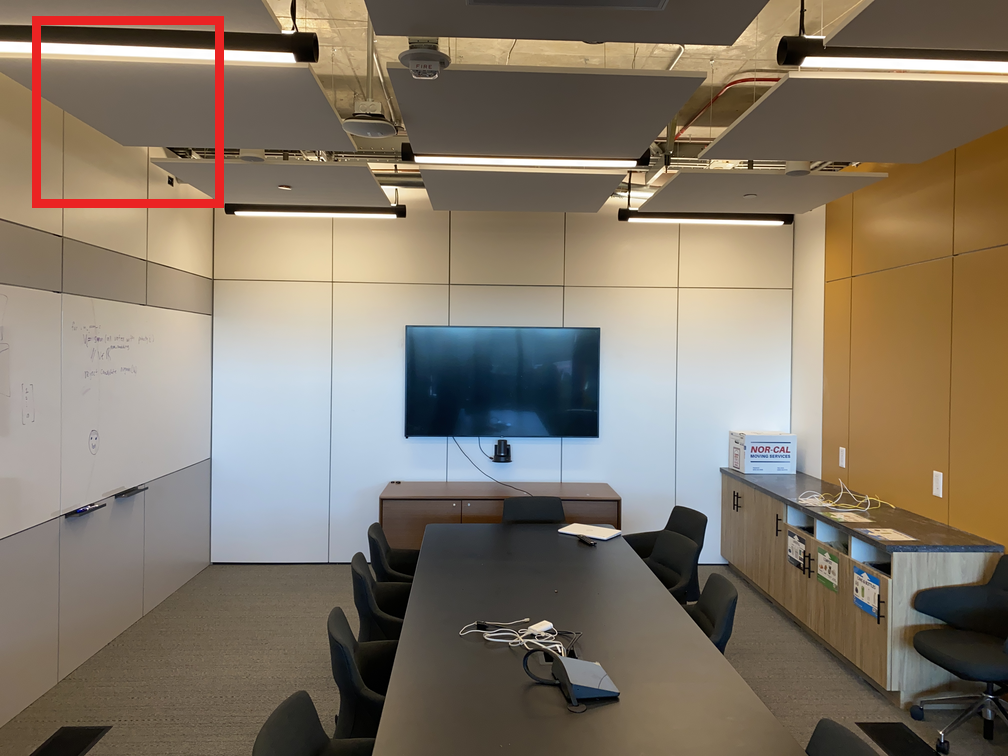} &
\includegraphics[width=0.127\textwidth]{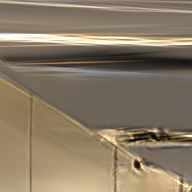} &
\includegraphics[width=0.127\textwidth]{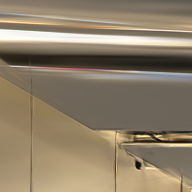} &
\includegraphics[width=0.127\textwidth]{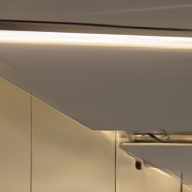} &
\includegraphics[width=0.127\textwidth]{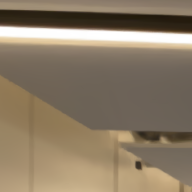} &
\includegraphics[width=0.127\textwidth]{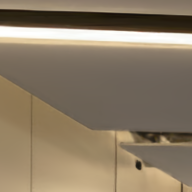} &
\includegraphics[width=0.127\textwidth]{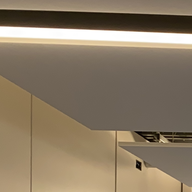}
\\

\includegraphics[width=0.168\textwidth]{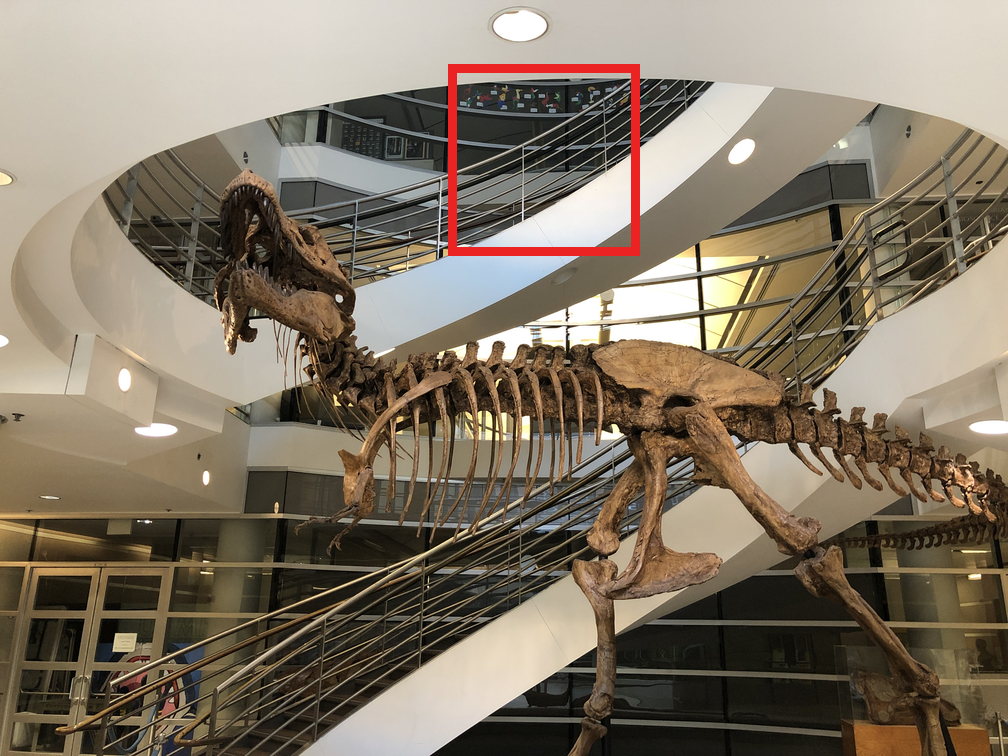} &
\includegraphics[width=0.127\textwidth]{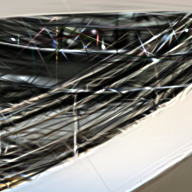} &
\includegraphics[width=0.127\textwidth]{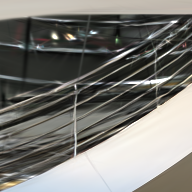} &
\includegraphics[width=0.127\textwidth]{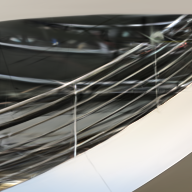} &
\includegraphics[width=0.127\textwidth]{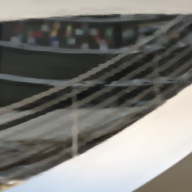} &
\includegraphics[width=0.127\textwidth]{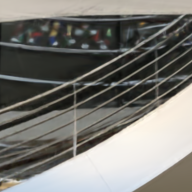} &
\includegraphics[width=0.127\textwidth]{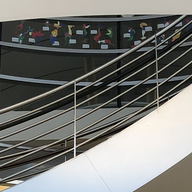}
\\
\multicolumn{1}{c}{\small Input} &
\multicolumn{1}{c}{\small 3DGS} &
\multicolumn{1}{c}{\small SRGS} &
\multicolumn{1}{c}{\makecell{\footnotesize DropGau.+SRGS}} &
\multicolumn{1}{c}{\small S2Gaussian} &
\multicolumn{1}{c}{\small CLEAR (Ours)} &
\multicolumn{1}{c}{\small GT}

\end{tabular}

\setlength{\abovecaptionskip}{5pt}
\caption{
Qualitative comparison on LLFF $4\times$ super-resolution.
Zoom in for the highlighted regions.
}
\label{fig:qualitative_LLFF}
\vspace{-5pt}
\end{figure*}

\begin{figure*}[t]
\vspace{-5pt}
\centering
\setlength{\tabcolsep}{1pt}

\begin{tabular}{cccccccc}
\multirow{2}{*}{%
\leavevmode\rotatebox{90}{\small\textbf{Blender}}%
}&
\includegraphics[width=0.122\textwidth]{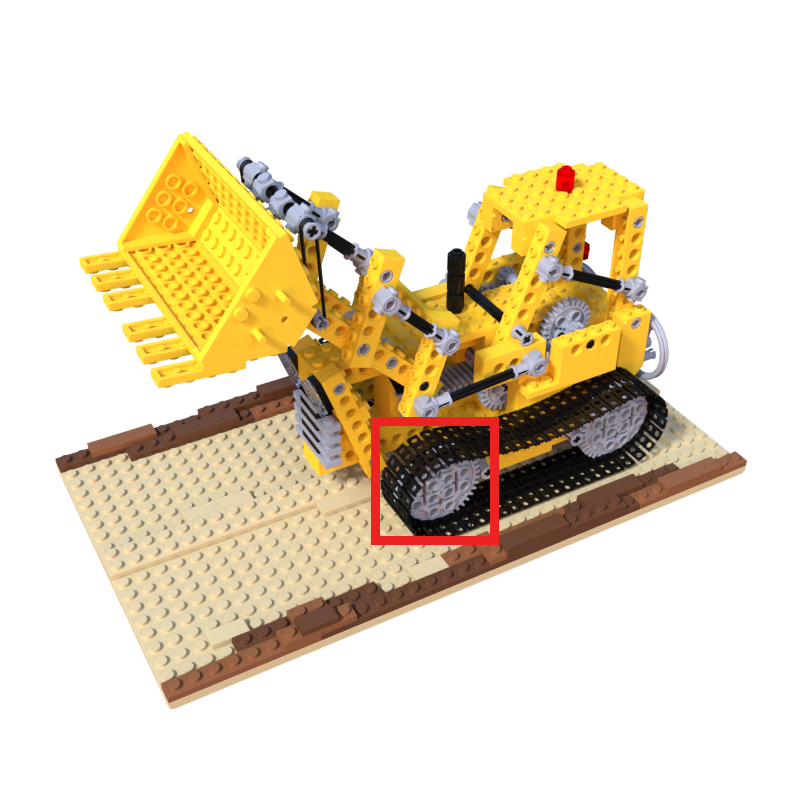} &
\includegraphics[width=0.122\textwidth]{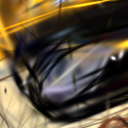} &
\includegraphics[width=0.122\textwidth]{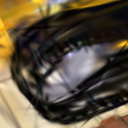} &
\includegraphics[width=0.122\textwidth]{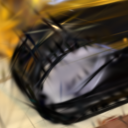} &
\includegraphics[width=0.122\textwidth]{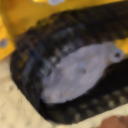} &
\includegraphics[width=0.122\textwidth]{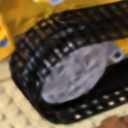} &
\includegraphics[width=0.122\textwidth]{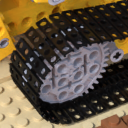}
\\
&
\includegraphics[width=0.122\textwidth]{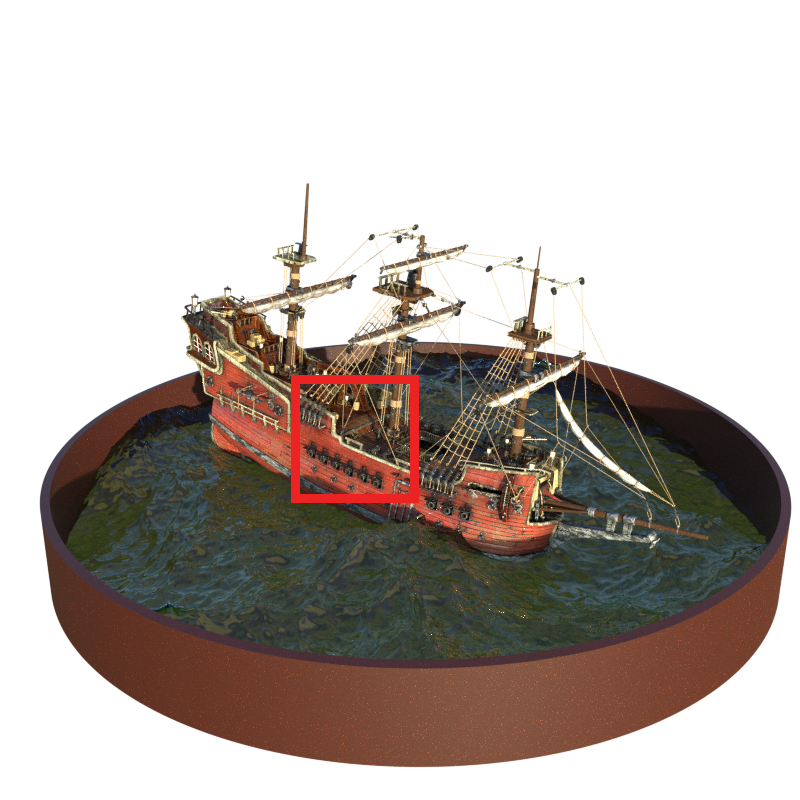} &
\includegraphics[width=0.122\textwidth]{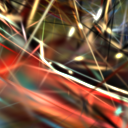} &
\includegraphics[width=0.122\textwidth]{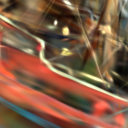} &
\includegraphics[width=0.122\textwidth]{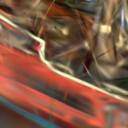} &
\includegraphics[width=0.122\textwidth]{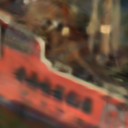} &
\includegraphics[width=0.122\textwidth]{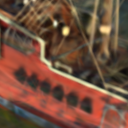} &
\includegraphics[width=0.122\textwidth]{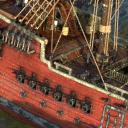}
\\

\multirow{2}{*}{\leavevmode\rotatebox{90}{\small\textbf{Mip-NeRF}}}
&
\includegraphics[width=0.183\textwidth]{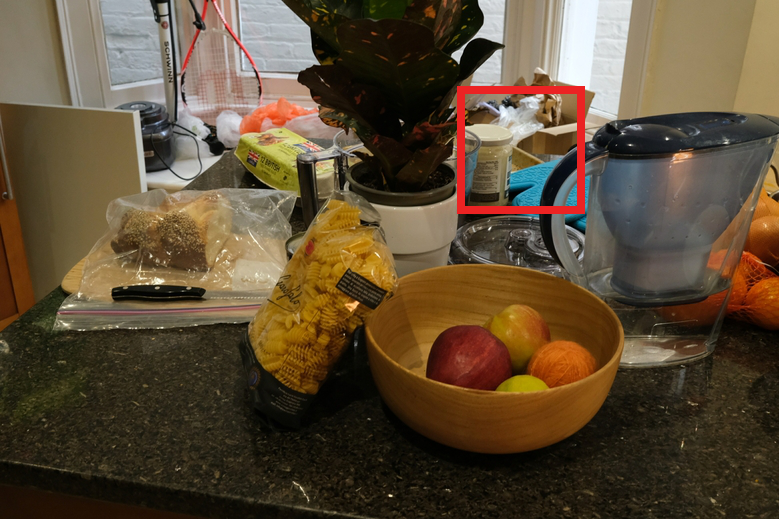} &
\includegraphics[width=0.122\textwidth]{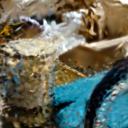} &
\includegraphics[width=0.122\textwidth]{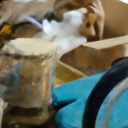} &
\includegraphics[width=0.122\textwidth]{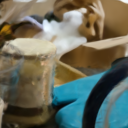} &
\includegraphics[width=0.122\textwidth]{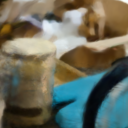} &
\includegraphics[width=0.122\textwidth]{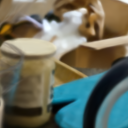} &
\includegraphics[width=0.122\textwidth]{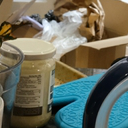} \\

 &
\includegraphics[width=0.183\textwidth]{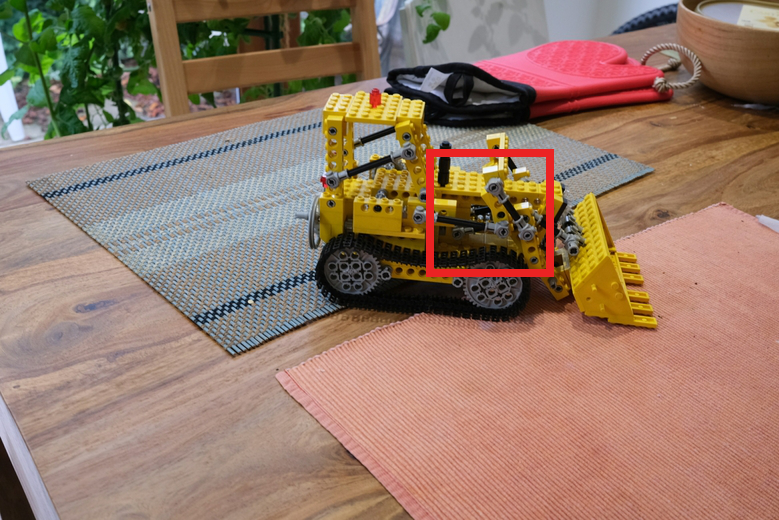} &
\includegraphics[width=0.122\textwidth]{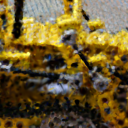} &
\includegraphics[width=0.122\textwidth]{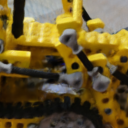} &
\includegraphics[width=0.122\textwidth]{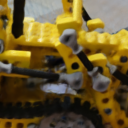} &
\includegraphics[width=0.122\textwidth]{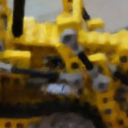} &
\includegraphics[width=0.122\textwidth]{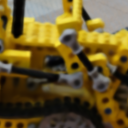} &
\includegraphics[width=0.122\textwidth]{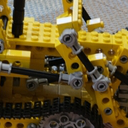} \\

&
\multicolumn{1}{c}{\small Input} &
\multicolumn{1}{c}{\small 3DGS} &
\multicolumn{1}{c}{\small SRGS} &
\multicolumn{1}{c}{\makecell{\footnotesize DropGau.+SRGS}} &
\multicolumn{1}{c}{\small S2Gaussian} &
\multicolumn{1}{c}{\small CLEAR (Ours)} &
\multicolumn{1}{c}{\small GT}

\end{tabular}
\setlength{\abovecaptionskip}{5pt}

\caption{
Qualitative comparison on Blender and  Mip-NeRF 360 $4\times$ super-resolution.
Zoom in for the highlighted regions.
}
\label{fig:qualitative_Blender_Mip}
\end{figure*}




\subsection{Sparse-View Stabilization}
\label{sec:stabilization}

Sparse-view optimization is prone to Gaussian co-adaptation and late-stage
geometric drift. To mitigate this, we first apply the same dropout mask to both rendering
scales:
\begin{equation}
\begin{aligned}
z_i^t
&=
\frac{\operatorname{Bernoulli}(1-p_t)}
     {1-p_t},\\
\widehat I_r^v
&=
\mathcal R_r
\left(
\bm z^t\odot\mathcal G_\theta,C^v
\right),
\quad
r\in\{\mathrm{LR},\mathrm{HR}\},
\end{aligned}
\label{eq:shared_gaussian_dropout}
\end{equation}
where $t$ is the current iteration, $z_i^t$ is the inverted-dropout mask of
Gaussian $i$, $p_t$ is the dropout rate, $\bm z^t$ collects all Gaussian
masks, $\odot$ denotes Gaussian masking, and $\widehat I_r^v$ is the rendering
of view $v$ at scale $r$. Sharing $\bm z^t$ prevents artificial LR and HR
inconsistency.

Once the Gaussian structure stabilizes at iteration $t_a$, we stop
densification and introduce a detached anchor:
\begin{equation}
\begin{aligned}
\theta^a
&=
\operatorname{sg}\!\left(\theta(t_a)\right),\\
\mathcal L_{\mathrm{stab}}
&=
\sum_i
\left\|
\theta_i-\theta_i^a
\right\|_{\bm W}^{2},
\qquad t\geq t_a ,
\end{aligned}
\label{eq:detached_anchor}
\end{equation}
where $t_a$ is the anchor iteration, $\theta^a$ is the detached Gaussian
state, $\theta_i^a$ is the anchor of Gaussian $i$,
$\operatorname{sg}(\cdot)$ denotes stop-gradient, $\mathcal L_{\mathrm{stab}}$ is the attribute-weighted anchor loss, and $\bm W$ controls
attribute-wise regularization. With this stabilization design, the geometric drift is suppressed while preserving limited appearance refinement.

\subsection{Training and Inference}
\label{sec:training_inference}

At each iteration, CLEAR first renders the shared Gaussian field at both the LR and HR scales. It computes patch-level evidence and lifts it to visible Gaussians, yielding the routed HR gradient $\bm g_i^{\mathrm{HR}}$. This is compared with the LR gradient $\bm g_i^{\mathrm{LR}}$ with destructive components corrected via Eq.~\eqref{eq:soft_conflict_correction}. The corrected gradients are merged via Eq.~\eqref{eq:conflict_merged_gradient}, and for $t\geq t_a$, the stabilization gradient $\lambda_{\mathrm{stab}} \nabla_{\theta_i}\mathcal L_{\mathrm{stab}}$ is further incorporated before the single optimizer update. Shared Gaussian dropout is applied to both rendering scales throughout training. During inference, all training-only regularization is disabled, and the optimized Gaussian field is rendered directly at the target HR resolution without stage-wise transfer or refinement.

\section{Experiments}

\begin{table}[t]
\centering

\setlength{\tabcolsep}{5.5pt}
\resizebox{\columnwidth}{!}{
\begin{tabular}{lccc}
\toprule
Method
& PSNR$\uparrow$
& SSIM$\uparrow$
& LPIPS$\downarrow$ \\
\midrule
\multicolumn{4}{l}{\textit{NeRF-based methods}} \\
NeRF-SR ~\cite{wang2022nerfsr}          & 9.28 & 0.226 & 0.617 \\
RegNeRF ~\cite{regnerf}                & 15.78 & 0.432 & 0.448 \\
\midrule
\multicolumn{4}{l}{\textit{3DGS-based methods}} \\
3DGS ~\cite{3dgs}                 & 12.51 & 0.325 & 0.573 \\
SRGS ~\cite{srgs}                 & 17.46 & 0.499 & 0.387 \\
Mip-Splatting ~\cite{mipsplatting}             & 15.05 & 0.456 & 0.517 \\
DropGaussian  ~\cite{dropgaussian}              & 18.57 & 0.551 & 0.390 \\
SplatSuRe ~\cite{splatsure}                       & 18.72 & 0.573 & 0.377 \\
DropGaussian ~\cite{dropgaussian}     \\ + SRGS~\cite{srgs}              & 19.14 & 0.555& 0.326 \\
DropGaussian ~\cite{dropgaussian} \\ + SplatSuRe ~\cite{splatsure}   & 19.33 & 0.578 & 0.333 \\
S2Gaussian$^{\dagger}$ ~\cite{s2gaussian}         & \underline{20.17} & \underline{0.640} & \underline{0.314}\\
\textbf{CLEAR (Ours)}       & \textbf{20.42} & \textbf{0.682} & \textbf{0.266} \\
\bottomrule
\end{tabular}
}
\caption{
Quantitative comparison on  LLFF $\times4$ with 3 input views.
The best and second-best results are highlighted in
\textbf{bold} and \underline{underline}, respectively. $\dagger$ denotes results reproduced by us.
}\label{tab:llff_results}
\vspace{-8pt}

\end{table}

\begin{table}[t]
\centering
\vspace{-5pt}
\setlength{\tabcolsep}{5.5pt}
\resizebox{\columnwidth}{!}{
\begin{tabular}{lccc}
\toprule
Method
& PSNR$\uparrow$
& SSIM$\uparrow$
& LPIPS$\downarrow$ \\
\midrule
\multicolumn{4}{l}{\textit{NeRF-based methods}} \\
NeRF-SR ~\cite{wang2022nerfsr} & 12.41 & 0.744 & 0.515 \\
RegNeRF ~\cite{regnerf} & 20.68 & 0.841 & 0.129 \\
\midrule
\multicolumn{4}{l}{\textit{3DGS-based methods}} \\
3DGS ~\cite{3dgs}            & 20.49 & 0.829 & 0.155 \\
SRGS ~\cite{srgs} & 22.47 & 0.852 & 0.126 \\
Mip-Splatting~\cite{mipsplatting}  & 22.66 & 0.859 & 0.121 \\
DropGaussian ~\cite{dropgaussian}               & 22.92 & 0.866 & 0.128 \\
SplatSuRe ~\cite{splatsure}                  & 23.01 & 0.867 & 0.119 \\
DropGaussian~\cite{dropgaussian}\\ + SRGS ~\cite{srgs}        & 22.99 & 0.866 & 0.119 \\
DropGaussian~\cite{dropgaussian}\\ + SplatSuRe ~\cite{splatsure}   & 23.09 & 0.872 & 0.117 \\
S2Gaussian$^{\dagger}$  ~\cite{s2gaussian}         & \underline{23.78} & \underline{0.876} & \underline{0.104} \\
\textbf{CLEAR (Ours)}       & \textbf{24.26} & \textbf{0.880} & \textbf{0.087} \\
\bottomrule
\end{tabular}
}
\caption{
Quantitative comparison on Blender $\times4$ with 8 input views.
The best and second-best results are highlighted in \textbf{bold} and \underline{underline}, respectively.
}\label{tab:blender_results}
\vspace{-7pt}
\end{table}

\begin{table}[t]
\centering
\vspace{-5pt}
\setlength{\tabcolsep}{5.5pt}
\resizebox{\columnwidth}{!}{
\begin{tabular}{lccc}
\toprule
Method
& PSNR$\uparrow$
& SSIM$\uparrow$
& LPIPS$\downarrow$ \\
\midrule
\multicolumn{4}{l}{\textit{NeRF-based methods}} \\
NeRF-SR ~\cite{wang2022nerfsr}                     & 10.26 & 0.269 & 0.628 \\
RegNeRF ~\cite{regnerf}                    & 17.28 & 0.417 & 0.449 \\
\midrule
\multicolumn{4}{l}{\textit{3DGS-based methods}} \\
3DGS~\cite{3dgs}                        & 16.69 & 0.357 & 0.488 \\
SRGS~\cite{srgs}                        & 18.52 & 0.466 & 0.389 \\
Mip-Splatting ~\cite{mipsplatting}              & 18.38 & 0.470 & 0.409 \\
DropGaussian ~\cite{dropgaussian}               & 18.74 & 0.489 & 0.400\ \\
SplatSuRe ~\cite{splatsure}                  & 19.02 & 0.501 & 0.389 \\
DropGaussian~\cite{dropgaussian}\\ + SRGS~\cite{srgs}         & 20.79 & 0.597 & 0.304 \\
DropGaussian~\cite{dropgaussian}\\ + SplatSuRe~\cite{splatsure}    & 21.12 & 0.624 & 0.302\\
S2Gaussian$^{\dagger}$~\cite{s2gaussian}         & \underline{21.96} & \underline{0.669} & \underline{0.298} \\
\textbf{CLEAR (Ours)}       & \textbf{22.19} & \textbf{0.697} & \textbf{0.288} \\
\bottomrule
\end{tabular}
}
\caption{
Quantitative comparison on  Mip-NeRF 360 $\times4$ with 24 input views.
The best and second-best results are highlighted in \textbf{bold} and \underline{underline}, respectively.
}\label{tab:Mip_results}
\vspace{-10pt}

\end{table}

\subsection{Implementation Details}
\label{sec:implementation}

\paragraph{Training Details.}
We optimize CLEAR for 10K iterations using Adam on a single NVIDIA A800 GPU. We adopt the default learning-rate configuration of Gaussian parameters in 3DGS~\cite{3dgs}. HR references are generated using the pre-trained ResShift model~\cite{resshift}. \textbf{For unified single-stage optimization}, we set $\lambda_{\mathrm{HR}}=1.0$ with a linear warm-up from iteration 500 to 3K and decay after 5K. 
We use $16\times16$ LR patches and corresponding $64\times64$ HR patches. For Gaussian-wise Conflict-aware Optimization \textbf{(GCO)}, we set $\tau_c=0.05$ with the correction scale factor bounded within $[0.05,0.75]$. For Evidence-Guided Patch-to-Gaussian Routing \textbf{(P2G)}, we set $\tau_r=0.05$, $q_{\min}=0.25$, and $\lambda_{\mathrm{HF}}=0.2$. For Sparse-View Stabilization \textbf{(SVS)}, the maximum Gaussian dropout rate is 0.2, and densification is performed every 100 iterations from 500 to 5K, and detached anchor regularization $\lambda_{\mathrm{stab}}=0.01$. The Mip filter scale and rasterization kernel are set to 0.2 and 0.1, respectively.

\paragraph{Datasets and Metrics.}
Following previous method, our experiments are conducted on three datasets under $\times4$ super-resolution: two real-world datasets, i.e., LLFF~\cite{mildenhall2019llff} and MipNeRF-360~\cite{mipnerf360}, and one synthetic dataset, i.e., Blender~\cite{nerf}. 
Moreover, we employ three evaluation metrics, i.e., PSNR, SSIM, and LPIPS,  for evaluating the rendering quality. 

\vspace{-2pt}
\paragraph{State-of-the-art Methods.}
We collect a range of representative methods, including 3DGS~\cite{3dgs}, Mip-Splatting~\cite{mipsplatting}, DropGaussian~\cite{dropgaussian}, SRGS~\cite{srgs}, and SplatSuRe~\cite{splatsure} as well as two-stage methods such as S2Gaussian~\cite{s2gaussian} and variants formed by combining DropGaussian with SRGS or SplatSuRe. We reproduce S2Gaussian due to its unreleased code. 

\subsection{Quantitative and Qualitative Comparisons}
\label{sec:comparisons}

\paragraph{Quantitative Evaluation.}
We evaluate CLEAR on three $4\times$ sparse-view super-resolution benchmarks: LLFF 3-views, Blender 8-views, and Mip-NeRF360 24-views, with results shown in Tabs.~\ref{tab:llff_results}-\ref{tab:Mip_results}, respectively.  We also summarize the PSNR and LPIPS performance on the three datasets in Fig.~\ref{fig:motivation}(c). These results consistently show that CLEAR achieves highest reconstruction quality on all three datasets, confirming its robustness to varying scene types and view sparsity levels. 
Moreover, we report the average training time of representative two-stage method S2Gaussian and our CLEAR on LLFF dataset in Tab.~\ref{tab:training_time}. The significantly higher training efficiency confirms the advantage of our unified single-stage framework.
These outstanding results in both performance and efficiency indicate that our unified single-stage optimization effectively reconciles reliable LR supervision with informative HR guidance, enabling more accurate and efficient HR Gaussian scene reconstruction. More results are in the supplementary material.

\begin{table}[t]
\centering
\setlength{\tabcolsep}{10pt}
\small
\begin{tabular}{@{}lcc@{}}
\toprule
Method &  S2Gaussian$^{\dagger}$ & \textbf{CLEAR (Ours)} \\
\midrule
LLFF (3 views) & 1028.25 & \textbf{612.37}\\
\bottomrule
\end{tabular}
\caption{
Average training time (s) on LLFF $4\times$ with 3 input views.
The reported time is averaged over all scenes.
$\dagger$ denotes results reproduced by us.
}
\label{tab:training_time}
\vspace{-5pt}
\end{table}

\paragraph{Qualitative  Evaluation.}
We further present visual comparisons for three datasets in
Fig.~\ref{fig:qualitative_LLFF} and Fig.~\ref{fig:qualitative_Blender_Mip}, respectively. Compared with previous approaches, CLEAR preserves better scene geometry under sparse-view supervision and reconstructs the better visual results. %
Notably, CLEAR exhibits accurate textural details on foliage and railings for LLFF, well-defined object structures for Blender, and significantly reduced artifacts for Mip-NeRF 360.
Collectively, these results substantiate that CLEAR effectively reconciles the structural cues from LR observations with the high-frequency information from SR guidance within a unified Gaussian representation, achieving both geometrically faithful and visually compelling high-resolution scene reconstruction.

\begin{table}[t]
\centering
\setlength{\tabcolsep}{9.0pt}
\small
\begin{tabular}{@{}lcccc@{}}
\toprule
Variant
& PSNR$\uparrow$
& SSIM$\uparrow$
& LPIPS$\downarrow$
& Conflict$\downarrow$ \\
\midrule
Baseline
& 18.74 & 0.569 & 0.375 & 0.243 \\

$+$ GCO
& 19.72 & 0.604 & 0.303 & 0.228 \\

$+$ P2G
& 20.09 & 0.629 & 0.287 & 0.210 \\

$+$ SVS (full)
& \textbf{20.42}
& \textbf{0.682}
& \textbf{0.266}
& \textbf{0.162} \\
\bottomrule

\end{tabular}
\caption{
Ablation studies on LLFF $4\times$ (3 views). 
The components are added progressively.
Conflict denotes the Gaussian-wise LR-HR gradient conflict ratio averaged over training.
}\label{tab:ablation}
\end{table}

\subsection{Ablation Studies}

\label{sec:ablation}


We conduct ablation studies on LLFF $4\times$ with three input views.
Starting from a single-stage baseline that directly combines LR and HR
supervision, we progressively introduce the designed GCO, P2G, and SVS.
The corresponding average results over all eight scenes are reported in Tab.~\ref{tab:ablation}.





\begin{figure}[!t]
\centering
\includegraphics[width=1\linewidth]
{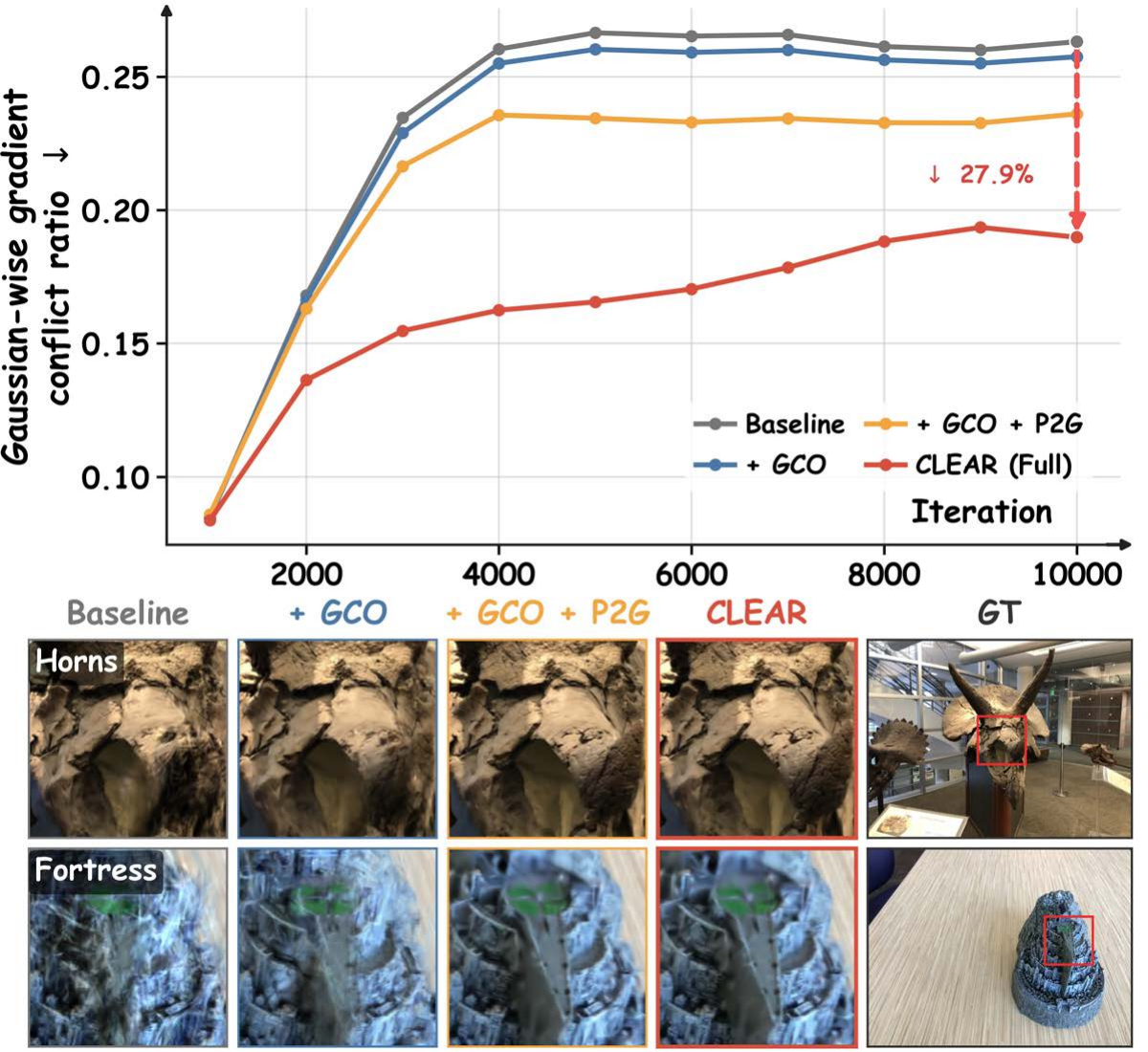}
\caption{
\textbf{Incremental analysis of CLEAR.}
\textbf{Top:} Gaussian-wise LR and HR gradient conflict ratio averaged over eight LLFF scenes during training.
\textbf{Bottom:} Novel-view results of different ablation models.
Our components progressively reduce gradient conflicts and improve reconstruction quality.
}
\label{fig:motivation_conflict}
\end{figure}

\noindent\textbf{Effect of GCO.} 
As shown in Tab.~\ref{tab:ablation}, GCO provides a effective balance between preserving authentic LR structure and leveraging HR supervision, achieved by correcting unreliable HR updates while preserving compatible gradients.

\noindent\textbf{Effect of P2G.}  P2G achieves a 0.37dB PSNR gain and reduces gradient conflicts from Tab.~\ref{tab:ablation}, which demonstrates that evidence-guided routing effectively concentrates high-frequency optimization and complements conflict correction, enabling more perceptually faithful reconstruction. 

\noindent\textbf{Effect of SVS.} SVS regularizes the unified Gaussian field against sparse-view overfitting, delivering a 0.33 dB PSNR gain and notably reducing gradient conflicts. Combined with GCO and P2G, it completes the CLEAR framework, attaining superior quality with minimal conflicts.

Furthermore, we provide the incremental analysis of CLEAR during training on LLFF dataset in Fig.~\ref{fig:motivation_conflict}. With increasing iterations, the gradient conflict steadily decreases, with each component contributing to this reduction and the reconstruction quality improves progressively.

\section{Conclusion}
\label{sec:conclusion}

We propose \textbf{CLEAR}, the first unified single-stage framework for Sparse-view 3D Gaussian Splatting Super-resolution. CLEAR performs joint the optimization of authentic LR observations and external HR priors within a unified single Gaussian field.    
A Gaussian-wise conflict-aware optimization strategy is proposed that reconciles inconsistent LR and HR gradients within each Gaussian. 
Moreover, an evidence-guided Patch-to-Gaussian routing mechanism is introduced that selectively propagates reliable super-resolving evidence to guide Gaussian densification.  
Additionally, a shared Gaussian dropout and a detached mid-training anchoring are adopted to enhance the robustness. Experiments on three $4\times$ super-resolution benchmarks demonstrate that CLEAR achieves superior rendering quality and geometric fidelity.

\newpage
\clearpage

\bibliography{aaai2027}

\end{document}